\documentclass[graybox]{svmult}

\smartqed  
\usepackage{mathptmx}       
\usepackage{helvet}         
\usepackage{courier}        
\usepackage{type1cm}        
\usepackage{graphicx}        
\usepackage{multicol}        
\usepackage[bottom]{footmisc}

\usepackage{amssymb}
\usepackage{amsmath}
\newcommand{\ind}{\mbox{1\hspace{-.25em}l}}

\newcommand{\betP}{\mathrm{BetP}}
\newcommand{\conj}{\mathrm{Conj}}
\newcommand{\dis}{\mathrm{Dis}}
\newcommand{\mix}{\mathrm{Mix}}

\newcommand{\DP}{\mathrm{DP}}
\newcommand{\Y}{\mathrm{Y}}
\newcommand{\Mean}{\mathrm{Mean}}
\newcommand{\DS}{\mathrm{DS}}
\newcommand{\caut}{\mathrm{Cautious}}
\newcommand{\PCR}{\mathrm{PCR6}}
\newcommand{\Flo}{\mathrm{Flo}}
\newcommand{\LNS}{\mathrm{LNS}}
\newcommand{\bel}{\mathrm{bel}}
\newcommand{\pl}{\mathrm{pl}}

\newcommand{\Conf}{\mathrm{Conf}}

\newcommand{\ocap}{\operatornamewithlimits{\text{\textcircled{\scalebox{1.4}{\tiny{$\cap$}}}}}}
\newcommand{\argmax}{\operatornamewithlimits{argmax}}

\begin{document}

\title*{Conflict management in information fusion with belief functions}
\author{Arnaud Martin}
\institute{A. Martin \at
              IRISA, University of Rennes 1
              \email{Arnaud.Martin@univ-rennes1.fr}}
%
%
\maketitle

\abstract*{In Information fusion, the conflict is an important concept. Indeed, combining several imperfect experts or sources allows conflict. In the theory of belief functions, this notion has been discussed a lot. The mass appearing on the empty set during the conjunctive combination rule is generally considered as conflict, but that is not really a conflict. Some measures of conflict have been proposed and some approaches have been proposed in order to manage this conflict or to decide with conflicting mass functions. We recall in this chapter some of them and we propose a discussion to consider the conflict in information fusion with the theory of belief functions.}

\abstract{In Information fusion, the conflict is an important concept. Indeed, combining several imperfect experts or sources allows conflict. In the theory of belief functions, this notion has been discussed a lot. The mass appearing on the empty set during the conjunctive combination rule is generally considered as conflict, but that is not really a conflict. Some measures of conflict have been proposed and some approaches have been proposed in order to manage this conflict or to decide with conflicting mass functions. We recall in this chapter some of them and we propose a discussion to consider the conflict in information fusion with the theory of belief functions.}

\section{Introduction}
\label{intro}
The theory of belief functions was first introduced by \cite{Dempster67} in order to represent some imprecise probabilities with {\em upper} and {\em lower probabilities}. Then \cite{Shafer76} proposed a mathematical theory of evidence with is now widely used for information fusion. Combining imperfect sources of information leads inevitably to conflict. One can consider that the conflict comes from the non-reliability of the sources or the sources do not give information on the same observation. In this last case, one must not combine them.

Let $\Omega=\{\omega_1,\ldots, \omega_n\}$ be a frame of discernment of exclusive and exhaustive hypothesis. A mass function $m$, also called basic belief assignment (bba) is the mapping from elements of the power set $2^\Omega$ (composed by all the disjunctions of $\Omega$) onto $[0,1]$ such that:
\begin{equation}
\sum_{X\in 2^\Omega} m(X)=1.
\label{hyp_norm}
\end{equation}
A focal element $X$ is an element of $2^\Omega$ such that $m(X)\neq 0$. If the focal elements are nested, the mass functions is {\em consonant}. A simple mass function, noted $A^w$ is given by:
\begin{equation}
\left\{
\begin{array}{l}
m(A)=w\\
m(\Omega)=1-w
\end{array}
\right.
\end{equation}
This mass function allows to show that we can model an imprecise information (if $A$ is an union of singletons $\omega_i$)  and an uncertain information (if $w>0$). All non-dogmatic mass functions (with $m(\Omega)>0)$) can be decomposed by a set of simple mass functions~\cite{Shafer76}.

Constraining $m(\emptyset)=0$ corresponds to a closed-world assumption \cite{Shafer76}, while allowing $m(\emptyset)\geq 0$ corresponds to an open world assumption \cite{Smets90}. Smets interpreted this mass on the empty set such as an non-expected hypothesis and normalizes it in the pignistic probability defined for all $X \in 2^\Omega$, with $X \neq \emptyset$ by:
\begin{eqnarray}
\label{pignistic}
\betP(X)=\sum_{Y \in 2^\Omega, Y \neq \emptyset} \frac{|X \cap Y|}{|Y|} \frac{m(Y)}{1-m(\emptyset)}.
\end{eqnarray}
The pignistic probability can be used in the decision process such as a compromise between the credibility and the plausibility. The credibility is given for all $X \in 2^\Omega$ by:
\begin{eqnarray}
\label{bel}
\bel(X)=\sum_{Y \subseteq X, Y \neq \emptyset} m(Y),
\end{eqnarray}
The plausibility is given for all $X \in 2^\Omega$ by:

\begin{eqnarray}
\label{pl}
\pl(X)=\sum_{Y \in 2^\Omega, Y\cap X \neq \emptyset} m(Y)=\bel(\Omega)-\bel(X^c)=1-m(\emptyset)-\bel(X^c),
\end{eqnarray}
where $X^c$ is the complementary of $X$. Hence, if we note the decision function $f_d$ that can be the pignistic probability, the credibility or the plausibility, we choose the element $A \in 2^\Omega$ for a given observation if:
\begin{eqnarray}
A=\argmax_{X \in \Omega} \left(f_d(X)\right).
\end{eqnarray}
The decision is made on the mass function obtained by the combination of all the mass function from the sources.

The first combination rule has been proposed by Dempster \cite{Dempster67} and is defined for two mass functions $m_1$ and $m_2$, for all $X \in 2^\Omega$, with $X\neq \emptyset$ by:
\begin{eqnarray}
m_\DS(X)=\displaystyle \frac{1}{1-\kappa}\sum_{A\cap B =X} m_1(A)m_2(B), 
\end{eqnarray}
where $\kappa= \displaystyle \sum_{A\cap B =\emptyset} m_1(A)m_2(B)$ is the inconsistence of the combination and generally called conflict. We call it here the {\em global conflict} such as the sum of all {\em partial conflicts}.

To stay in an open world, Smets \cite{Smets90} proposes the non-normalized conjunctive rule given for two mass functions $m_1$ and $m_2$ and for all $X \in 2^\Omega$ by:
\begin{eqnarray}
\label{conjunctive}
m_\conj(X)=\displaystyle \sum_{A\cap B =X} m_1(A)m_2(B):=(m_1\ocap m_2 )(X).
\end{eqnarray}

These both rules allow to reduce the imprecision of the focal elements and to increase the belief on concordant elements after the fusion. The main assumptions to apply these rules are the cognitive independence and the reliability of the sources.   

Based on the results of these rules, the problem enlightened by the famous Zadeh's example \cite{Zadeh84} is the repartition of the global conflict. Indeed, consider \linebreak $\Omega=\{\omega_1,\omega_2,\omega_3\}$ and two experts opinions given by $m_1(\omega_1)=0.9$, $m_1(\omega_3)=0.1$, and $m_2(\omega_2)=0.9$, $m_1(\omega_3)=0.1$, the mass function resulting in the combination using Dempster's rule is $m(\omega_3)=1$ and using conjunctive rule is $m(\emptyset)=0.99$, \linebreak $m(\omega_3)=0.01$. Therefore, several combination rules have been proposed to manage this global conflict \cite{Smets07,Martin07}. 


As observed in \cite{Liu06,Martin08}, the weight of conflict given by $\kappa=m_\conj(\emptyset)$ is not a conflict measure between the mass functions. Indeed, the conjunctive-based rules are not idempotent (as the majority of the rules defined to manage the global conflict): the combination of identical mass functions leads generally to a positive value of $\kappa$. Hence, new kind of conflict measures are defined in \cite{Martin08}.

In the following section~\ref{sec:2}, we recall some measures of conflict in the theory of belief functions. Then, in section~\ref{sec:3} we present the ways to manage the conflict either before the combination, or in the combination rule. The last section~\ref{sec:4} presents some decision methods in order to consider the conflict during this last step of information process.

\section{Modeling conflict}
\label{sec:2}

First of all, we should not mix up conflict measure and contradiction measure. The measures defined by \cite{George96,Wierman01} are not conflict measures, but some discord and specificity measures (to take the terms of \cite{Klir94}) we call contradiction measures. We define the contradiction and conflict measures by the following definitions:

\textbf{Definition} \textit{A {\em contradiction} in the theory of belief functions quantifies how a mass function contradicts itself.}
 
\textbf{Definition} (C1) \textit{The {\em conflict} in the theory of belief functions can be defined by the contradiction between two or more mass functions.}

Therefore, is the mass of the empty set or the functions of this mass (such as $-\ln(1-m_\conj(\emptyset))$ proposed by \cite{Shafer76}) a conflict measure? It seems obvious that the property of the non-idempotence is a problem to use this as a conflict measure. However, if we define a conflict measure such as $\Conf(m_1,m_2)=m_\conj(\emptyset)$, we note that $\Conf(m_1,m_\Omega)=0$ where $m_\Omega(\Omega)=1$ is the ignorance. Indeed, the ignorance is the neutral element for the conjunctive combination rule. This property seems to be followed from a conflict measure.

Other conflict measures have been defined. In \cite{Jousselme11}, a conflict measure is given by: 
\begin{equation}
 \Conf(m_1,m_2)=1-\frac{\textbf{pl}_1^T.\textbf{pl}_2}{\|\textbf{pl}_1\|\|\textbf{pl}_2\|}
\end{equation}
where $\textbf{pl}$ is the plausibility function and $\textbf{pl}_1^T.\textbf{pl}_2$ the vector product in $2^n$ space of both plausibility functions. However, generally $\Conf(m_1,m_\Omega)\neq 0$, that seems counter-intuitive.

\paragraph{\bf Auto-conflict}\quad

Introduced by \cite{Osswald06}, the auto-conflict of order $s$ for one source is given by: 
\begin{eqnarray}
\label{autoconf}
a_s=\displaystyle \left(\mathop{\ocap}_{j=1}^{^{\mbox{\small $s$}}} m\right)(\emptyset).
\end{eqnarray}
where $\mathop{\ocap}$ is the conjunctive operator of Equation~\eqref{conjunctive}. The following property holds: $a_s\leq a_{s+1}$, 
meaning that due to the non-indempotence of $\mathop{\ocap}$, the more masses $m$ are combined with itself the nearer to 1 $\kappa$ is, and so in a general case, the more the number of sources is high the nearer to 1 $\kappa$ is. The behavior of the auto-conflict was studied in \cite{Martin08} and it was shown that we should take into account the auto-conflict in the global conflict in order to really define a conflict. In \cite{Yager92}, the auto-conflict was defined and called the plausibility of the belief structure with itself. The auto-conflict is a kind of measure of the contradiction, but depends on the order $s$ of the combination. A measure of contradiction independent on the order has been defined in \cite{Smarandache11}.

\paragraph{\bf Conflict measure based on a distance}\quad

With the definition of the conflict (C1), we consider sources to be in conflict if their opinions are far from each other in the space of corresponding bba's. That suggests a notion of distance. That is the reason why in \cite{Martin08}, we give a definition of the measure of conflict between sources assertions through a distance between their respective bba's. The conflict measure between $2$ experts is defined by:
\begin{eqnarray}
\label{2conflict_measure}
\Conf(1,2)=d(m_1,m_2).
\end{eqnarray}
We defined the conflict measure between one source $j$ and the other $M-1$ sources by:
\begin{eqnarray}
\label{conflict_measure1}
\Conf(j,\mathcal{E})=\frac{1}{M-1}\sum_{i=1, i\neq j}^M \Conf(i,j),
\end{eqnarray}
where $\mathcal{E}=\{1,\ldots, M\}$ is the set of sources in conflict with $j$. Another definition is given by:
\begin{eqnarray}
\label{conflict_measure2}
\Conf(j,M)=d(m_j,\overline{m_M}),
\end{eqnarray}
where $\overline{m_M}$ is the bba of the artificial source representing the combined opinions of all the sources in $\mathcal{E}$ except $j$.

A comparison of distances in the theory of belief functions is presented in \cite{Jousselme11}. We consider the distance defined in \cite{Jousselme01} as the most appropriate. This distance is defined for two basic belief assignments $\textbf{m}_1$ and $\textbf{m}_2$ on $2^\Omega$ by:
\begin{eqnarray}
\label{distance}
d_J(m_1,m_2)=\sqrt{\frac{1}{2} (\textbf{m}_1-\textbf{m}_2)^T\underline{\underline{D}}(\textbf{m}_1-\textbf{m}_2)},
\end{eqnarray}
where $\underline{\underline{D}}$ is an $2^{|\Omega|}\times 2^{|\Omega|}$ matrix based on Jaccard dissimilarity whose elements are:
\begin{eqnarray}
\label{DMatrix}
D(A,B)=\left\{
\begin{array}{l}
1, \, \mbox{if} \, A= B=\emptyset,\\
\\
\displaystyle \frac{|A\cap B|}{|A\cup B|}, \, \forall A, B \in 2^\Omega.\\
\end{array}
\right.
\end{eqnarray}

An interesting property of this measure is given by $\Conf(m,m)=0$. That means that there is no conflict between a source and itself (that is not a contradiction). However, we generally do not have $\Conf(m,m_\Omega)=0$, where $m_\Omega(\Omega)=1$ is the ignorance.

\paragraph{\bf Conflict measure based on inclusion degree and distance}\quad
\label{conflict}

We have seen that we cannot use the mass on the empty set as a conflict measure because of the non-idempotence of the conjunctive rule. We also have seen that the conflict measure based on the distance is not null in general for the ignorance mass. The conjunctive rule does not transfer mass on the empty set if the mass functions are {\em included}. We give here some definitions of the inclusion.

\textbf{Definition 1: strict inclusion\\} \textit{We say that the mass function $m_1$ is {\em included} in $m_2$ if all the focal elements of $m_1$ are included in \textbf{each} focal elements of $m_2$.}

\textbf{Definition 2: light inclusion\\} \textit{We say that the mass function $m_1$ is {\em included} in $m_2$ if all the focal elements of $m_1$ are included in \textbf{at least one} focal element of $m_2$.} 

\textbf{Definition} \textit{We note this inclusion by $m_1 \subseteq m_2$. The mass functions are {\em included} if $m_1$ is included in $m_2$ or $m_2$ is included in $m_1$.}


In~\cite{Martin12}, we propose a conflict measure base on five following axioms. Let note $\Conf(m_1,m_2)$ a conflict measure between the mass functions $m_1$ and $m_2$. We present hereafter essential properties that must verify a conflict measure.

\begin{enumerate}
 \item Non-negativity: 
\begin{equation}
\Conf(m_1,m_2)\geq 0
\end{equation}

A negative conflict does not make sense. This axiom is,therefore, necessary.

\item Identity: 
\begin{equation}
\Conf(m_1,m_1)= 0 
\end{equation}

Two equal mass functions are not in conflict. This property is not reached by the global conflict, but seems natural.

\item Symmetry: 
\begin{equation}
\Conf(m_1,m_2)= \Conf(m_2,m_1) 
\end{equation}

The conflict measure must be symmetric. We do not see any case where the non-symmetry can make sense.

\item Normalization: 
\begin{equation}
0 \leq \Conf(m_1,m_2) \leq 1 
\end{equation}

This axiom is may not be necessary to define a conflict measure, but the normalization is very useful in many applications requiring a conflict measure.

\item Inclusion: 
\begin{equation}
\Conf(m_1,m_2) = 0, \mbox{ if and only if } m_1 \subseteq m_2 \mbox{ or } m_2 \subseteq m_1
\end{equation}

This axiom means that if the focal elements of two mass functions are not conflicting (the intersection is never empty), the mass functions are not in conflict and the mass functions cannot be in conflict if they are included. This axiom is not satisfied by a distance based conflict measure. 
\end{enumerate}
These proposed axioms are very similar to ones defined in~\cite{Desterke12}. If a conflict measure satisfied these axioms that is not necessary a distance. Indeed, we only impose the identity and not the definiteness ($\Conf(m_1,m_2) = 0 \Leftrightarrow m_1=m_2$). \linebreak The axiom of inclusion is less restrictive and makes more senss for a conflict measure. Moreover, we do not impose the triangle inequality \linebreak ($\Conf(m_1,m_2) \leq \Conf(m_1,m_3)+\Conf(m_3,m_2)$). It can be interesting to have \linebreak $\Conf(m_1,m_2) \geq \Conf(m_1,m_3)+\Conf(m_3,m_2)$ meaning that an expert given the mass function $m_3$ can reduce the conflict. Therefore, a distance (with the property of the triangle inequality) cannot be used directly to define a conflict measure.


We see that the axiom of inclusion seems very important to define a conflict measure. This is the reason why we define in~\cite{Martin12} a degree of inclusion to measure how two mass functions are included. Let the inclusion index: $Inc(X_1, Y_2) = 1$ if $X_1 \subseteq Y_2$ and 0 otherwise, where $X_1$ and $Y_2$ are two focal elements of $m_1$ and $m_2$ respectively. According to the definition 1 and definition 2, we introduce two degrees of inclusion of $m_1$ {\bf in} $m_2$. A strict degree of inclusion  of $m_1$ {\bf in} $m_2$ is given by:
\begin{equation}
 d_{incS}(m_1,m_2)= \frac{1}{|{\cal F}_1||{\cal F}_2|} \sum_{X_1 \in {\cal F}_1} \sum_{ Y_2 \in {\cal F}_2} Inc(X_1, Y_2)
\end{equation}
where ${\cal F}_1$ and ${\cal F}_2$ are the set of focal elements of $m_1$ and $m_2$ respectively, and $|{\cal F}_1|$, $|{\cal F}_2|$ are the number of focal elements of $m_1$ and $m_2$. 

This definition is very strict, so we introduce a light degree of inclusion  of $m_1$ {\bf in} $m_2$ given by:
\begin{equation}
 d_{incL}(m_1,m_2)= \frac{1}{|{\cal F}_1|} \sum_{X_1 \in {\cal F}_1} \max_{ Y_2 \in {\cal F}_2} Inc(X_1, Y_2).
\end{equation}

%

Let $\delta_{inc}(m_1,m_2)$ a degree of inclusion of $m_1$ {\bf and} $m_2$ define by:
\begin{equation}
 \delta_{inc}(m_1,m_2)=\max(d_{inc}(m_1,m_2), d_{inc}(m_2,m_1))
\end{equation}
This degree gives the maximum of the proportion of focal elements from one mass function included in another one. Therefore, $\delta_{inc}(m_1,m_2)=1$ if and only if $m_1$ and $m_2$ are included, and the axiom of inclusion is reached for $1-\delta_{inc}(m_1,m_2)$.

Hence, we define in~\cite{Martin12}, a conflict measure between two mass functions $m_1$ and $m_2$ by:
\begin{equation}
 \Conf(m_1,m_2)=(1-\delta_{inc}(m_1,m_2)) d_J(m_1,m_2)
\end{equation}
where $d_J$ is the distance defined by the equation~\eqref{distance}. All the previous axioms are satisfied. Indeed the axiom of inclusion is $1-\delta_{inc}(m_1,m_2)$ and the distance $d_J$ satisfied the other axioms. Moreover $0 \leq \delta_{inc}(m_1,m_2) \leq 1$, by the product of $1-\delta_{inc}$ and $d_J$, all the axioms are satisfied.

For more than two mass functions, the conflict measure between one source $j$ and the other $M-1$ sources can be defined from equations~\eqref{conflict_measure1} or \eqref{conflict_measure2}.



\section{Managing conflict}
\label{sec:3}

The role of conflict is essential in information fusion. Different ways can be used to manage and reduce the conflict. The conflict can come from the low reliability of the sources. Therefore, we can use this conflict to estimate the reliability of the sources if we cannot learn it on databases as proposed in \cite{Martin08}. Hence, we reduce the conflict before the combination, but we can also directly manage the conflict in the rule of combination as generally made in the theory of belief functions such as explained in \cite{Smets07,Martin07}. 

According to the application, we do not search always to reduce the conflict. For example, we can use the conflict measure such as an indicator of the inconsistence of the fusion for example in databases \cite{Chebbah10}. Conflict information can also be an interesting information in some applications such as presented in \cite{Rominger10}. 

\subsection{Managing the conflict before the combination}

The conflict appearing while confronting several experts' opinions can be used as an indicator of the relative reliability of the experts. We have seen that there exist many rules in order to take into account the conflict during the combination step. These rules do not make the difference between the conflict (global or local conflict) and the auto-conflict due to the non-idempotence of the majority of the rules. We propose here the use of a conflict measure in order to define a reliability measure, that we consider before the combination, in a discounting procedure.

When we can quantify the reliability of each source, we can weaken the basic belief assignment before the combination by the discounting procedure:
 \begin{eqnarray}
 \label{discounting}
 \left\{
 \begin{array}{l}
 	m_j^\alpha(X)=\alpha_j m_j(X), \, \forall X \in 2^\Omega \smallsetminus \{\Omega\} \\
 	m_j^\alpha(\Omega)=1-\alpha_j (1-m_j(\Omega)).
 \end{array}
 \right.
 \end{eqnarray}
$\alpha_j \in [0,1]$ is the discounting factor of the source $j$ that is, in this case, the reliability of the source $j$, eventually as a function of $X \in 2^\Omega$.

Other discounting procedures are possible such as the contextual discounting \cite{Mercier06}, or a discounting procedure based on the credibility or the plausibility functions \cite{Zeng07}.

According to the applications, we can learn the discounting factors $\alpha_j$, for example, from the confusion matrix \cite{Martin05}. In many applications, we cannot learn the reliability of each source. A general approach to the evaluation of the discounting factor without learning is given in \cite{Elouedi04}. For a given bba the discounting factor is obtained by the minimization on $\alpha$ of a distance given by:
\begin{eqnarray}
\label{reliability_Smets}
Dist^{\alpha_j}=\sum_{A\in \Omega} \left(\betP_j(A)-\delta_{A,j}\right)^2,
\end{eqnarray}
where $\betP_j$ is the pignistic probability (Equation~\eqref{pignistic}) of the bba given by the source $j$ and $\delta_{A,j}=1$ if the source $j$ supports $A$ and 0 otherwise.

This approach is interesting with the goal of making decision based on pignistic probabilities. However, if the source $j$ does not support a singleton of $\Omega$, the minimization on $\alpha_j$ does not work well.

\begin{figure}[htb]
\begin{center}
\includegraphics[height=5cm]{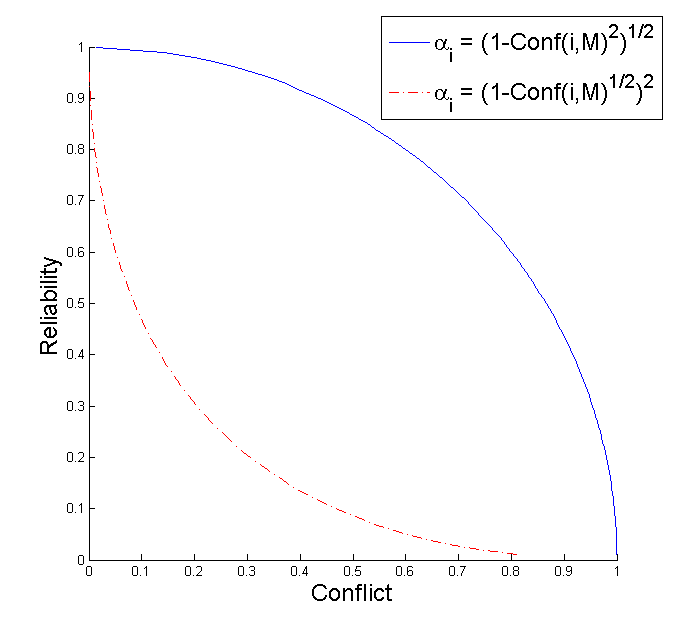}
\end{center}
\caption{Reliability of one source based on conflict of the source with the other sources.}
\label{Fig_reliability}
\end{figure}

In order to combine the bbas of all sources together, we propose here to estimate the reliability of each source $j$ from the conflict measure $Conf$ between the source $j$ and the others  by:
\begin{eqnarray}
\alpha_j=f(\Conf(j,M)),
\end{eqnarray}
where $f$ is a decreasing function. We can choose:
\begin{eqnarray}
\label{reliability}
\alpha_j=\left(1-\Conf(j,M)^\lambda\right)^{1/\lambda},
\end{eqnarray}
where $\lambda>0$. We illustrate this function for $\lambda=2$ and \linebreak $\lambda=1/2$ on figure~\ref{Fig_reliability}. This function allows to give more reliability to the sources with few conflict with the other. 

Other definitions are possible. The credibility degree defined in \cite{Chen05} is also based on the distance given in the Equation~\eqref{distance}, and could also be interpreted as the reliability of the source. However the credibility degree in~\cite{Chen05} is integrated directly in the combination with a weighted average. Our reliability measure allows the use of all the existing combination rules.

\subsection{Managing the conflict in the combination}
According to the application, if we cannot reduce the conflict before the combination, we can do it by incorporating it into a combination rule. The choice of the combination rule is not easy, but it can be done by utilizing the conflict and the assumption on its origin. Indeed the Dempster rule can be apply if the sources are independent and reliable. Dempster's rule is given for $S$ sources for all $X \in 2^\Omega$, $X\neq \emptyset$ by:
\begin{eqnarray}
\label{DS}
m_\DS(X)=\frac{1}{1-m_\conj(\emptyset)} \sum_{Y_1 \cap \ldots \cap Y_s = X}  \prod_{j=1}^S m_j(Y_j)=\frac{m_\conj(X)}{1-\kappa},
\end{eqnarray}
where $\kappa=m_\conj(\emptyset)$, $Y_j \in 2^\Omega$ is the answer of the source $S_j$, and $m_j(Y_j)$ the associated mass function. This normalization by $1-m_\conj(\emptyset)$ hide the conflict and so this rule is interesting only if we consider the closed-world and if the sources are not highly conflicting. 

If the assumption of independent and reliable sources are not reached, the application of the Dempster's rule can produce some global conflict.

\paragraph{\bf Conflict coming from a false assumption of closed world}\quad

In the closed world, the elements of the frame of discernment are assumed exhaustive. If $m(\emptyset)> 0$, this mass can be interpreted such as another element, and so the assumption on the exhaustiveness of the frame of discernment is false. 
Hence, Smets~\cite{Smets07} proposed the use of the conjunctive rule given for $S$ sources for all $X \in 2^\Omega$ by:
\begin{eqnarray}
\label{conj}
m_\conj(X)=\sum_{Y_1 \cap \ldots \cap Y_s = X}  \prod_{j=1}^S m_j(Y_j).
\end{eqnarray}
Here, the sources must be cognitively independent and reliable while the open-world is considered. Hence, the mass on the empty set can be interpreted as another element unknown by the sources. In fact, in the proposed model by Smets, the conflict is transfered during the decision step by the pignistic probability~\cite{Smets90}, hiding the conflict to the end. This rule cannot be used in applications with a high value of $\kappa$.

The global conflict come from the sum of the partial conflict. Hence, if the global conflict can be interpreted as an unknown element, all the partial conflict can be interpreted as many unknown elements. In that case we can keep the partial conflict in order to decide on these elements (see section~\ref{sec:4} for this consideration). Therefore the assumption of the exclusivity is considered here as false. 

Under this assumption, the mass functions are no more defined on the power set $2^\Omega$ but on the so called hyper power set\footnote{This notation is introduced by~\cite{Dezert02a} and $D$ come from the Dedeking lattice.} $D^\Omega$. Therefore the space $\Omega$ is closed by the union and intersection operators. This extension of the power set, lead to a lot of reflexions around this new expressiveness taking the name of DSmT (\emph{Dempster-Shafer modified Theory}). 

One can also consider a partial exclusiveness of the frame of discernment. Hence, we introduce the notation $D_r^\Omega$ in~\cite{Martin09a} in order to integrate some constraints on the exclusiveness of some elements of $\Omega$ and so reduce the hyper power set size. Under these assumptions, we define the $\PCR$ rule in~\cite{Martin06a}, given by:
\begin{eqnarray}
\label{PCR6combination}
\begin{array}{rcl}
  \displaystyle m_\PCR(X)  &=&  \displaystyle m_\conj(X) \\
  \!\!\!\!\!\!\!\!\!\!\!\!\!&+& \displaystyle \sum_{j=1}^S   m_j(X)^2 
  \!\!\!\!\!\!\!\! \displaystyle \sum_{\begin{array}{c}
      \scriptstyle {\displaystyle \mathop{\cap}_{j^\prime=1}^{S\!-\!1}} Y_{\sigma_j(j^\prime)} \cap X = \emptyset \\
      \scriptstyle (Y_{\sigma_j(1)},\ldots ,Y_{\sigma_j(S\!-\!1)})\in (D_r^\Omega)^{s\!-\!1}
  \end{array}}
  \!\!\!\!\!\!\!\!\!\!\!\!
  \left(\!\!\frac{\displaystyle \prod_{j^\prime=1}^{s\!-\!1} m_{\sigma_j(j^\prime)}(Y_{\sigma_j(j^\prime)})}
       {\displaystyle m_j(X) \!+\! \sum_{j^\prime=1}^{s\!-\!1} m_{\sigma_j(j^\prime)}(Y_{\sigma_j(j^\prime)})}\!\!\right)\!\!,
\end{array}
\end{eqnarray}
where $\sigma$ is given by:
\begin{eqnarray}
\label{sigma}
\left\{
\begin{array}{ll}
\sigma_j(j^\prime)=j^\prime &\mbox{if~} j^\prime<j,\\
\sigma_j(j^\prime)=j^\prime+1 &\mbox{if~} j^\prime\geq j,\\
\end{array}
\right.
\end{eqnarray}
This rule transfers the partial conflicts on the elements that generate it, proportionally to their masses. This rule has been used in many applications allowing for good results.

\paragraph{\bf Conflict coming from the assumption of source's independence}\quad

If we consider some dependent sources, the conjunctive rule cannot be used. If we want to combine mass functions coming from dependent sources, the combination rule has to be idempotent. Indeed, if we combine two identical dependent mass functions (coming from different dependent sources), we have to obtain the same mass function without any global conflict. 

The simplest way to obtain a non-idempotent rule is the average of the mass functions such given in~\cite{Murphy00} by:
\begin{eqnarray}
\label{Murphy}
m_\Mean(X)=\displaystyle\frac{1}{s} \sum_{j=1}^S m_j(Y_j).
\end{eqnarray}
We showed the interest in a such rule in~\cite{Osswald06}, but in that case the sources have to be assumed totally reliable. If the sources give high conflicting information, this rule can provide some errors in the decision process.

The cautious rule \cite{Denoeux08} could be used to combine mass functions for which independence assumption is not verified. Cautious combination of $S$ non-dogmatic mass functions $m_j, j=1,2,\cdots,S$  is defined by the bba with the following weight function:
\begin{equation}
  w(A)= \mathop{\wedge}\limits_{j=1}^S w_j(A), ~~ A \in 2^\Omega \setminus \Omega.
\end{equation}
We thus have
\begin{equation}
 m_\caut(X) = \ocap_{A \subsetneq \Omega} A^{\mathop{\wedge}\limits_{j=1}^S w_j(A)},
\end{equation}
where $A^{w_j(A)}$ is the simple support function focused on $A$ with weight function $w_j(A)$ issued from the canonical decomposition of $m_j$. Note also that $\wedge$ is the min operator.

When the dependence/independence of the sources is estimated, another rule was proposed in~\cite{Chebbah15a}.

\paragraph{\bf Conflict coming from source's ignorance}\quad

Another possible interpretation of the reason for the conflict is the ignorance of the sources. Indeed, if a source is highly ignorant, it should give a categorical mass function on $\Omega$. 

Therefore, \cite{Yager87} interprets the global conflict coming from the ignorance of the sources and transfers the mass on the total ignorance ({\em i.e.} on $\Omega$) in order to keep the closed world assumption. In the case of high conflict, the result of the fusion is the ignorance. This rule is given by:
\begin{eqnarray}
\label{Yager}
\begin{array} {rcl}
m_\Y(X)&=& m_\conj(X), \forall X \in 2^\Omega, \, X\neq \emptyset, \, X \neq \Omega \\
m_\Y(\Omega)&=&m_\conj(\Omega)+m_\conj(\emptyset)\\
m_\Y(\emptyset)&=&0.
\end{array}
\end{eqnarray}

A source can also be ignorant not on all but only on some focal elements. Hence, \cite{Dubois88} proposed a clever conflict repartition by transferring the partial conflicts on the partial ignorances. This rule is given for all $X \in 2^\Omega$, $X\neq \emptyset$ by: 
\begin{eqnarray}
\label{DP}
m_\DP(X)=\sum_{Y_1 \cap \ldots \cap Y_s = X} \prod_{j=1}^S m_j(Y_j)+\sum_{
\scriptstyle{\begin{array}{c}
Y_1 \cup \ldots \cup Y_s = X\\
Y_1 \cap \ldots \cap Y_s = \emptyset \\
\end{array}}} \prod_{j=1}^S m_j(Y_j),
\end{eqnarray}
where $Y_j \in 2^\Omega$ is a focal element of the source $S_j$, and $m_j(Y_j)$ the associated mass function. This rule has a high memory complexity, such as the $\PCR$ rule, because it is necessary to manage the partial conflict.

\paragraph{\bf Conflict coming from source reliability assumption}\quad

If we have no knowledge of the reliability of the sources, but we know that at least on source is reliable, the disjunctive combination can be used. It is given for all $X \in 2^\Omega$ by:
\begin{eqnarray}
\label{disjunctive}
m_\dis(X)=\displaystyle\sum_{Y_1 \cup \ldots \cup Y_s = X}  \prod_{j=1}^S m_j(Y_j).
\end{eqnarray}
The main problem of this rule is the lost of specificity after combination.

One can also see the global conflict $\kappa=m_\conj(\emptyset)$ such as an estimation of the conflict coming from the unreliability of the sources. Therefore, the global conflict can play the role of a weight between a conjunctive and disjunctive comportment of the rule such introduced by~\cite{Florea09a}. This rule is given for $X \in 2^\Omega$, $X\neq \emptyset$ by:
\begin{eqnarray}
\label{Florea}
m_\Flo(X)=\displaystyle \beta_1(\kappa) m_\dis(X)+ \beta_2(\kappa) m_\conj(X),
\end{eqnarray}
where $\beta_1$ and $\beta_2$ can be defined by: 
\begin{eqnarray}
\begin{array}{l}
\beta_1(\kappa)=\displaystyle \frac{\kappa}{1-\kappa+\kappa^2},\\
\beta_2(\kappa) = \displaystyle \frac{1-\kappa}{1-\kappa+\kappa^2}.\\
\end{array}
\end{eqnarray}

In a more general way, we propose in~\cite{Martin07} to regulate the conjunctive/disjunctive comportment taking into consideration the partial combinations. The mixed rule is given for $m_1$ and $m_2$ for all $X \in 2^\Omega$ by:
\begin{eqnarray}
\label{mix}
\begin{array}{rcl}
m_\mix(X)&=&\displaystyle \sum_{Y_1\cup Y_2 =X} \delta_1 m_1(Y_1)m_2(Y_2)\\
&+& \displaystyle \sum_{Y_1\cap Y_2 =X} \delta_2 m_1(Y_1)m_2(Y_2).
\end{array}
\end{eqnarray}
If $ \delta_1=\beta_1(\kappa)$ and $\delta_2=\beta_2(\kappa)$ we obtain the rule of~\cite{Florea09a}. Likewise, if \linebreak $\delta_1=1- \delta_2=0$ we obtain the conjunctive rule, and if $\delta_1=1- \delta_2=1$ the disjunctive rule. With $\delta_1(Y_1,Y_2)=1- \delta_2(Y_1,Y_2)=\ind_{\{\emptyset\}}(Y_1\cap Y_2)$ we get back to the rule of~\cite{Dubois88} by taking into account partial conflicts.

The choice of $\delta_1=1- \delta_2$ can also be made from a dissimilarity such as: 
\begin{eqnarray}
\label{eq_delta2}
\delta_2(Y_1,Y_2)=\displaystyle \frac{|Y_1\cap Y_2|}{\min (|Y_1|, |Y_2|)},
\end{eqnarray}
where $|Y_1|$ is the cardinality of $Y_1$. Jaccard dissimilarity can also be considered by:
\begin{eqnarray}
\label{eq_d2}
\delta_2(Y_1,Y_2)=\displaystyle \frac{|Y_1\cap Y_2|}{|Y_1\cup Y_2|}.
\end{eqnarray}
Therefore, if we have a partial conflict between $Y_1$ and $Y_2$, $|Y_1\cap Y_2|=0$ and the rule transfers the mass on $Y_1\cup Y_2$. In that case $Y_1\subset Y_2$ (or the contrary), $Y_1\cap Y_2=Y_1$ and $Y_1 \cup Y_2=Y_2$, hence with $\delta_2$ given by~\eqref{eq_delta2} the rule transfers the mass on $Y_1$ and with $\delta_2$ given by~\eqref{eq_d2} on $Y_1$ and $Y_2$ according to the ratio $|Y_1|/|Y_2|$ of cardinalities. In the case $Y_1\cap Y_2 \neq Y_1, Y_2$ and $\emptyset$, the rule transfers the mass on $Y_1 \cap Y_2$ and $Y_1 \cup Y_2$ according to equations~\eqref{eq_delta2} and~\eqref{eq_d2}. With such weights, the $\mix$ rule considers partial conflict according to the imprecision of the elements at the origin of the conflict.

\paragraph{\bf Conflict coming from a number of sources}\quad

When we have to combine a many sources, the assumption of the reliability of all the sources is difficult to consider especially if the sources are human. The disjunctive rule~\eqref{disjunctive} assume that at least one source is reliable but a precise decision will be difficult to take. Moreover, the complexity of main rules managing the conflict in a clever way is too high such as the rules given by~\eqref{mix} and \eqref{PCR6combination}. That is the reason why we introduce in~\cite{Zhou17} a new rule according to the following assumptions:
\begin{itemize}
\item The majority of sources are reliable;
 \item The larger extent one source is consistent with others, the more reliable the source is;
 \item The sources are cognitively independent.
\end{itemize}

For each mass function $m_j$ we consider the set of mass functions $\{A_k^{w_j}, A_k \subset \Omega\}$ coming from the canonical decomposition. If group the simple mass functions $A_k^{w_j}$ in $c$ clusters (the number of distinct $A_k$) and denote by $s_k$ the number of simple mass functions in the cluster $k$, the proposed rule is given by:
  \begin{equation}
  \label{lastcombination}
   m_\LNS=\ocap_{k=1, \cdots, c} (A_k)^{1-\alpha_k+\alpha_k \displaystyle \prod_{j=1}^{s _k} w_j}
  \end{equation}
where 
\begin{equation}
      \label{discountfactorSimple}
      \alpha_k = \frac{s_k}{\displaystyle  \sum_{i=1}^{c} s_i}.
\end{equation}

\paragraph{\bf How to choose the combination rule?}\quad

To answer the delicate question on which combination rule to choose, many authors propose a new rule. Of course, we could propose a \emph{no free lunch theorem} showing that there is no a best combination rule. 

To answer this question, we propose in~\cite{Martin09b} a global approach to transfer the belief. Indeed, the discounting process, the reduction of the number of focal elements, the combination rules and the decision process can be seen such as a transfer of belief and we can define these transfers in joint operator. However, it seems to difficult to propose a global approach which will be too general to be applied. In~\cite{Martin09b}, we define a rule integrating only the reliability given for $X\in 2^\Omega$ by:
\begin{eqnarray}
\label{GenComb}
m(X)\!\!=\displaystyle \!\! \!\!\!
\sum_{{\bf Y} \in (2^\Omega)^S} \! \prod_{j=1}^S \! m_j(Y_j) w(X,{\bf m}({\bf Y}),{\cal T}({\bf Y}), {\bf \alpha}({\bf Y})),
\end{eqnarray}
where ${\bf Y}=(Y_1,\ldots , Y_S)$ is the response vector of the $S$ sources, $m_j(Y_j)$ the associated masses (${\bf m}({\bf Y})=(m_1(Y_1),\ldots,m_j(Y_s))$, $w$ is a weight function, ${\bf \alpha}$ is the matrix of terms $\alpha_{ij}$ of the reliability of the source $S_j$ for the element $i$ of $2^\Omega$, and ${\cal T}({\bf Y})$ is the set of elements of $2^\Omega$, on which we transfer the elementary masses $m_j(Y_j)$ for a given vector ${\bf Y}$. This rule has been illustrated in a special case integrating the local reliability, but it seems even too complex to be easily applied. 

Hence, the best way to choose the combination rule is to identify the assumptions that we can or we have to make and choose the adapted rule according to these assumptions. 

However, we know that a given rule can provide good results in a context where the assumptions are satisfy this rule. Hence, another way to evaluate and compare some rules of combination, is to study the results (after decision) of the combined masses, {\em e.g.} on generated mass functions. In~\cite{Osswald06}, from generated mass functions, we study the difference of the combination rules in terms of decisions. We showed that we have to take into account the decision process. We will present some of them in the next section in the context of conflicting mass functions.

\section{Decision with conflicting bbas}
\label{sec:4}

The classic functions used for decision making such as the pignistic probability, the credibility and the plausibility are increasing by sets inclusion. We cannot use these functions directly to decide on other elements than the singletons.
When the assumption of exclusiveness of the frame of discernment is not made, such as in the equation~\eqref{PCR6combination}, we can decide on $D_r^\Omega$. That can be interesting from the data mining point of view such as we show in~\cite{Martin08b}. 

The approach proposed by~\cite{Appriou14a} has been extend to the consideration of $D_r^\Omega$ in~\cite{Martin09a} allowing to decide on any element of $D_r^\Omega$ by taking into account the mass function and the cardinality. Hence, we choose the element $A \in D_r^\Omega$ for a given observation if:
\begin{eqnarray}
\label{DecAppriouDSmT}
	A=\argmax_{X \in D_r^\Omega} \left(m_d(X)f_d(X)\right),
\end{eqnarray}
where $f_d$ is the considered decision function such as the credibility, plausibility or pignistic probability calculated from the mass function coming from the result of the combination rule, and $m_d$ is the mass function defined by:
\begin{eqnarray}
\label{MasseBayes}
m_d(X)=K_d \lambda_X \left(\frac{1}{{\cal C_M}(X)^\rho}\right),
\end{eqnarray}
${\cal C_M}(X)$ is the cardinality of $X$ of $D_r^\Omega$, defined by the number of disjoint parts in the Venn diagram, $\rho$ is a parameter with its values in $[0,1]$ allowing to decide from the intersection of all the singletons (with $\rho=1$) until the total ignorance $\Omega$ (with $\rho=0$). The parameter $\lambda_X$ enables us to integrate the lost of knowledge on one of the elements $X$ of $D_r^\Omega$. The constant $K_d$ is a normalization factor that guaranties the condition of equation \eqref{hyp_norm}. Without any constraint on $D^\Omega$, all the focal elements contain the intersection of the singletons. One cannot choose the plausibility such as decision function $f_d$.

The choice of the parameter $\rho$ is not easy to make. It depends on the size of $\Omega$. According to the application, it can be more interesting to define a subset on which we want to take the decision. Hence, we can envisage the decision on any subset of $D_r^\Omega$, noted ${\cal D}$, and equation~\eqref{DecAppriouDSmT} becomes simply:
\begin{eqnarray}
\label{DecGen}
	A=\argmax_{X \in {\cal D}} \left(m_d(X)f_d(X)\right).
\end{eqnarray}
Particularly this subset can be defined according to the expected cardinality of the element on which we want to decide.

With the same spirit, in~\cite{Essaid14}, another decision process is proposed by:
\begin{eqnarray}
	A=\argmax_{X \in {\cal D}} \left(d_J(m,m_X)\right),
\end{eqnarray}
where $m_X$ is the categorical mass function $m(X)=1$, and $m$ is the mass function coming from the combination rule. The subset ${\cal D}$ is the set of elements on which we want to decide. 

This last decision process allows also a decision on imprecise elements of the power set $2^\Omega$ and to control the precision of expected decision element without any parameter to fit.

\section{Conclusion}
\label{conclusion}

In this chapter, we propose some solutions to the problem of the conflict in information fusion in the context of the theory of belief functions. In section~\ref{sec:2} we present some conflict measures. Today, there is no consensus in the community on the choice of a conflict measure. Measuring the conflict is not an easy task because a mass function contains some information such as auto-conflict, we can interpret differently. The proposed axioms are a minimum that a conflict measure has to reach. In section~\ref{sec:3}, we discuss how to manage the conflict. Based on the assumption that conflict comes from the unreliability of the sources, with a conflict estimation for each source, the best to do is to discount the mass function according to the reliability estimation (and so the conflict measure).

Another way to manage the conflict, is the choice of the combination rule. Starting from the famous Zadeh's example, many combination rules have been proposed to manage the conflict. In this chapter, we present some combination rules (without exhaustiveness) according to the assumptions that the rules suppose. Hence, we distinguish the assumptions of: open/closed world, dependent/independent sources, ignorant/not-ignorant sources, reliable/unreliable sources, few/many sources.

To end this chapter, when the assumption of exclusiveness of the frame of discernment is not make, and so when we postpone the matter of conflict to the decision, we present some adapted decision processes. These decision methods are also adapted to decide on some imprecise elements of the power set. 

Of course, all the exposed methods here must be selected according to the application, to the possible assumptions, and to the final expected result.

\end{document}